# A Fast-Converged Acoustic Modeling for Korean Speech Recognition: A Preliminary Study on Time Delay Neural Network

Hosung Park*, Donghyun Lee*, Minkyu Lim*, Yoseb Kang*, Juneseok Oh*
and Ji-Hwan Kim*

*Department of Computer Science and Engineering, Sogang University,
Seoul 04107, Republic of Korea
E-mail: hosungpark@sogang.ac.kr

**Abstract**

In this paper, a time delay neural network (TDNN) based acoustic model is proposed to implement a fast-converged acoustic modeling for Korean speech recognition. The TDNN has an advantage in fast-convergence where the amount of training data is limited, due to subsampling which excludes duplicated weights. The TDNN showed an absolute improvement of 2.12% in terms of character error rate compared to feed forward neural network (FFNN) based modelling for Korean speech corpora. The proposed model converged 1.67 times faster than a FFNN-based model did.
**Key Words**: Time Delay Neural Network, Deep Neural Network, Acoustic Model, Speech Recognition

## 1. Introduction

Speech recognition requires an acoustic model that can effectively learn the context of adjacent input speech features to improve the recognition performance [1]. General cognitive tasks are also characterized by a high degree of uncertainty and variability. For these tasks, it is difficult to achieve good performance using standard serial programming methods. In recent years, neural networks have been widely used to solve these problems. They have been applied to acoustic models in speech recognition.

However, neural networks such as FFNN and recurrent neural network (RNN), require large amount of speech data in convergence of model training, and as a result, it consumes a lot of time in model training. Whereas, a TDNN is a type of neural network which has independent relationship between the length of input context and number of sequence steps. A subsampling method is used to reduce duplicated weights between nodes in this neural network [1].

In this paper, a TDNN model is proposed and its performance is evaluated for Korean speech corpora. The remainder of this paper is organized as follows. In Section 2, related works are described. Section 3 presents the experiments. A conclusion is given in Section 4.

## 2. Related works

A convolutional neural network (CNN) was used to extraction of speech features in speech recognition [2]. In the research, A CNN-TDNN based architecture was proposed for the raw waveform setup. The raw waveform setup using CNN showed similar performance to the MFCC-based one [2, 3, 4].

RNN model was used in acoustic modeling to consider the wider range context of speech signal [5]. RNN was trained by considering the short context of input signal as well as long context using a long short-term memory (LSTM) structure. However, a lot of training time was required.

TDNN was applied to learn the context of short intervals between input features like mel-scale frequency cepstral coefficients (MFCC) and that of long intervals in upper layers [1,4]. Its lower layers learn short contexts, and higher layers do long contexts of input.

The TDNN structure is shown in Figure 1. A general FFNN learns entire input features for processing contexts [1]. However, the TDNN architecture is learned on narrow contexts and the upper layers of the networks process wider contexts of the input features. Each layer in a TDNN is updated by different resolution which increases in higher layers of the network.

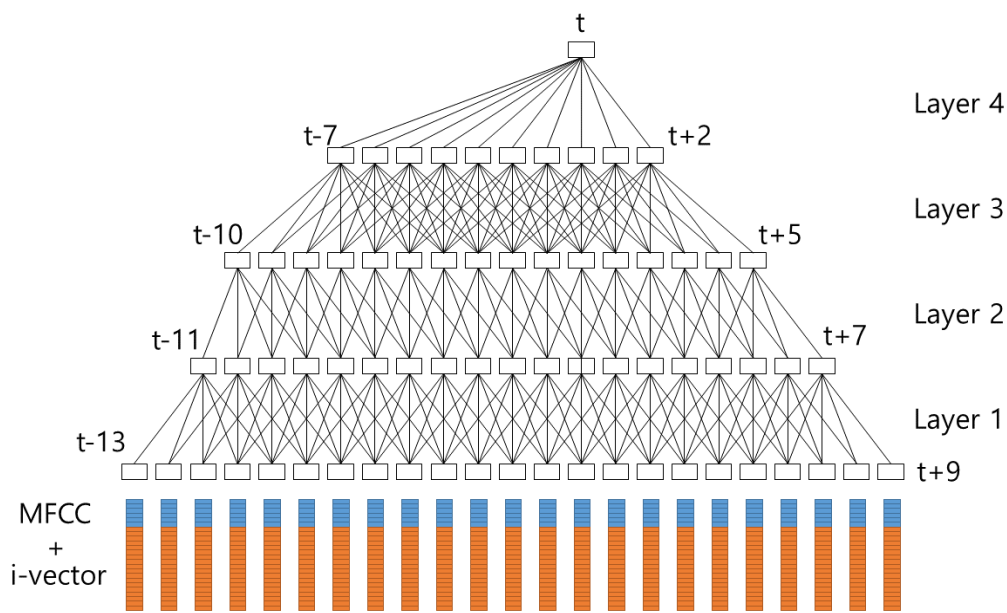

Fig. 1. TDNN architecture.

The TDNN is optimized by excluding duplicated weights in the networks. A standard FFNN learns overlapped features. If these duplicated updates are removed, the amount of time in training is reduced. A subsampling is used to reduce duplicated inputs. Figure 2 shows an example of TDNN architecture after subsampling is applied. Nodes and weights represented by dashed line are only updated when subsampling is applied. The method does not connect two or more inputs in hidden layer by allowing a space between frames [3]. If the interval between frames is allowed, the model can learn all input features because TDNN has a long context going up to the upper layer. Additionally, the number of parameters representing the model is reduced by minimizing the number of edges and nodes in the network.

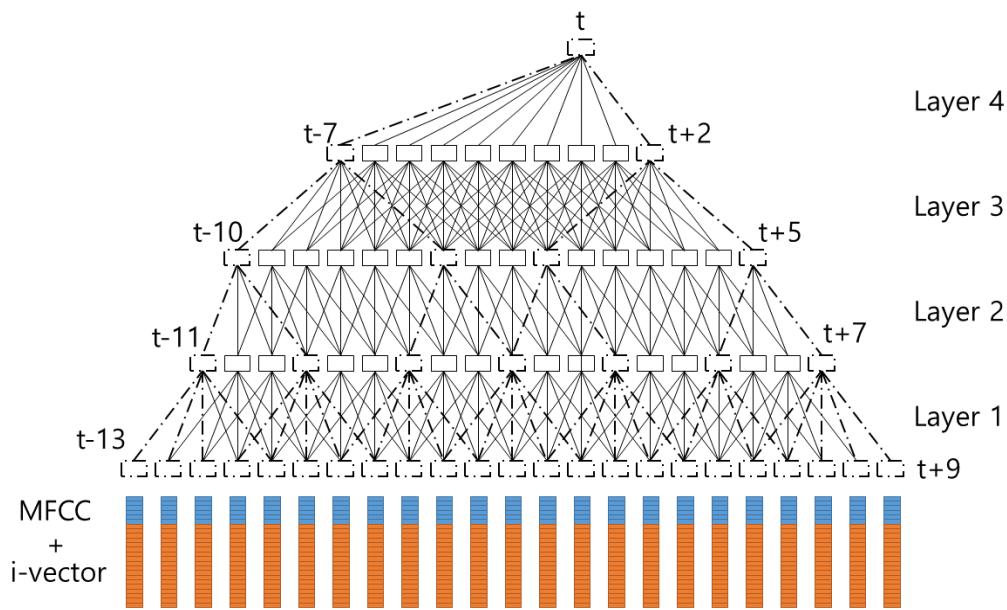

Fig. 2. An example of TDNN architecture after subsampling is applied

A time delay neural network architecture captures long term temporal dependencies better than a feed-forward DNN does [1]. In the research, a relative improvement of 7.3% was shown compared to the baseline DNN model in interactive personal assistant (IPA) domain. The TDNN was also applied in the removal of reverberation effect for robust acoustic modeling. iVectors were used as an input to the TDNN to perform instantaneous speaker and environment adaptation. Ten percent relative improvement was measured in terms of word error rate.

## 3. Experiments

Kaldi speech recognition toolkit was used to develop acoustic models [7]. The Kaldi is a C++ and python-based open source software toolkit that provides a variety of deep learning-based acoustic model architectures. In addition, it supports WFST-based decoding networks [7]. A 740 hours Korean corpus was used as training data. Two hours of test data, collected in the interactive personal assistance (IPA) domain, were used to evaluate the performance of speech recognition systems.

The speech corpus was recorded with the sampling rate of 16,000 samples per second, 16-bit resolution, and mono channel. 40 dimensional MFCC features were extracted from windows of 25ms length. In addition, 100 dimensional iVectors were added to the MFCC input. The iVector presents speaker characteristics The SRILM toolkit [8] was used to generate tri-gram language model using frequency cut.

The TDNN learned the way to update network parameters by using greedy layer-wise supervised training [9]. This method was used to train neural networks in Kaldi, which supports multiple GPUs in the training. In this paper, four GPUs were used to train the TDNN. An sMBR [10], which was a state-level variant of the minimum phone error (MPE), was used as an objective function in the training.

### 3.1 Results

Performance were compared between four neural net based acoustic models. Table 1 summarizes this comparison. TDNN-A and TDNN-B are TDNN models using the hyperparameters used in the Kaldi scripts released for Wall street journal corpus and Librispeech corpus, respectively [11,12]. TDNN-C uses 40 dimensional MFCCs and a 100 dimensional iVectors. Character error rate (CER) is used as an evaluation metric of speech recognition.

Table 1. Performance comparison between neural network architectures

| Acoustic model network type | No. of nodes in a hidden layer | No. of Hidden layers | CER(%) |
|---|---|---|---|
| FFNN | 3,600 | 4 | 3.58 |
| TDNN-A | 650 | 4 | 2.36 |
| **TDNN-B** | **1,280** | **5** | **1.97** |
| TDNN-C(iVector) | 1,280 | 5 | 2.17 |

The best CER was measured as 1.97% for TDNN-B. This CER is smaller by absolute 1.41% than that of FFNN. In addition, the TDNN-based acoustic model learns faster than FFNN. The TDNN-based acoustic model required only 60% of time in model training.

### 4. Conclusion

This paper showed the performance evaluation of the TDNN-based acoustic model for Korean corpora. TDNN has an advantage in fast-convergence when the amount of training data is limited, as subsampling excludes duplicated weights. It showed an absolute improvement of 2.12% in terms of CER compared to FFNN based model. In addition, the proposed model uses only 60% less training time than FFNN based model does.

## 5. Acknowledgments

This work was supported by the Technology Innovation Program (10080681, Technical development of Korean speech recognition system in vehicle) funded By the Ministry of Trade, Industry & Energy (MOTIE, Korea)